\documentclass[letterpaper, 10 pt, conference]{ieeeconf}  

\IEEEoverridecommandlockouts                              

\usepackage{graphics} 
\usepackage{graphicx}
\usepackage{mathptmx} 
\usepackage{amsmath} 
\usepackage{amssymb}  

\graphicspath{ {./Figures/} }

\usepackage{algorithm}
\usepackage{algorithmic}
\usepackage{multirow}
\usepackage{tabularx}

\title{\LARGE \bf
CASPNet++: Joint Multi-Agent Motion Prediction
}

\author{Maximilian Schäfer$^{1}$, Kun Zhao$^{2}$ and Anton Kummert$^{1}$
\thanks{$^{1}$Maximilian Schäfer and Anton Kummert are with the School of Electrical, Information and Media Engineering,
        University of Wuppertal, 42119 Wuppertal, Germany {\tt\small  \{maximilian.schaefer, kummert\}@uni-wuppertal.de}}%
\thanks{$^{2}$Kun Zhao is with Aptiv Services Deutschland GmbH
        \tt\small kun.zhao@aptiv.com}%
}

\begin{document}

\maketitle
\thispagestyle{empty}
\pagestyle{empty}

\begin{abstract}
	The prediction of road users' future motion is a critical task in supporting advanced driver-assistance systems (ADAS). It plays an even more crucial role for autonomous driving (AD) in enabling the planning and execution of safe driving maneuvers.
	Based on our previous work, Context-Aware Scene Prediction Network (CASPNet), an improved system, $\text{CASPNet++}$, is proposed. In this work, we focus on further enhancing the interaction modeling and scene understanding to support the joint prediction of all road users in a scene using spatiotemporal grids to model future occupancy. Moreover, an instance-based output head is introduced to provide multi-modal trajectories for agents of interest. In extensive quantitative and qualitative analysis, we demonstrate the scalability of CASPNet++ in utilizing and fusing diverse environmental input sources such as HD maps, Radar detection, and Lidar segmentation.
	Tested on the urban-focused prediction dataset nuScenes, CASPNet++ reaches state-of-the-art performance. The model has been deployed in a testing vehicle, running in real-time with moderate computational resources.
\end{abstract}

\section{INTRODUCTION}
The prediction of road users' future motion is a crucial task in the automotive field. A reliable prediction is vital to plan a safe and comfortable trajectory for the ego vehicle and to enable advanced driver assistance systems such as adaptive cruise control and automated lane change. It is a challenging task, especially when facing highly complex urban environments.

\begin{figure}[ht!]
	\centering
	\includegraphics[width=0.49\textwidth]{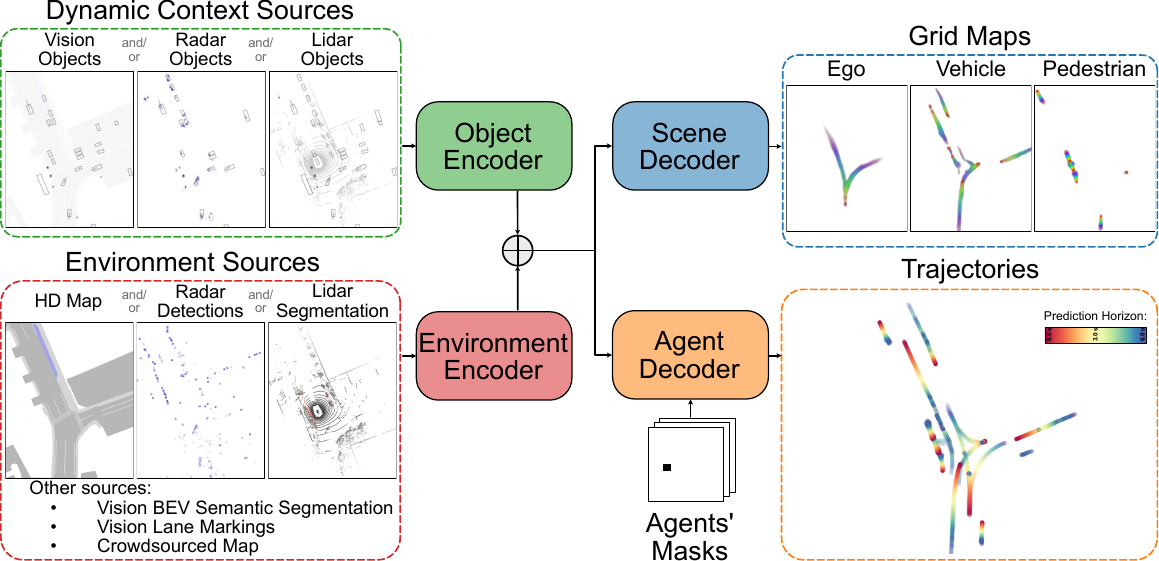}
	\caption{Model overview of CASPNet++. As inputs, the model can use diverse dynamic and environmental context sources. As output, a series of grid maps are predicted by the Scene Decoder to describe the future occupancy of all objects in a scene. Furthermore, the Agent Decoder predicts multi-modal trajectories for agents of interest.}
	\label{fig_architecture_overview}
\end{figure}

In this work, we propose Context-Aware Scene Prediction Network++ (CASPNet++), which takes on the task of urban-focused scene prediction utilizing grid-based input and output data representations. Spatiotemporal output grids enable the model to efficiently predict every road user in a scene while also facilitating the network to inherently learn diverse plausible futures. Regardless of the scenes' complexity, the proposed system maintains a constant computational cost due to its fixed-size input and output grids. This is an important advantage that is often neglected.
In addition, an Agent Decoder (Fig. \ref{fig_architecture_overview}) allows the model to predict multi-modal trajectories for agents of interest to support planning applications.

The environmental context of a scene in prediction benchmarks is often given in the form of an HD map \cite{Caesar2020,zhan2019interaction}. However, in practice, HD maps are expensive, not always accurate at times, and only available for a limited number of regions, restricting their real-world widespread application. At the same time, different ADAS systems may have various sensor setups. Rather than designing and developing for each setup separately, it is essential to have a system that can be adapted easily to different sensor setups and HD map availability. Fig. \ref{fig_architecture_overview} provides a general overview of our proposed system. 

Our contributions can be summarized as follows:
\begin{itemize}
	\item We propose a novel network architecture to jointly predict all road users in a scene while at the same time enabling efficient instance-based agent predictions.
	\item We demonstrate that our model can be easily adapted to utilize diverse environmental context sources, in order to support various ADAS/AD setups.
	\item At the same time, efficiency and runtime are key considerations in the system design. CASPNet++ does not only reach state-of-the-art results on nuScenes \cite{Caesar2020} but is also tested in our testing vehicle in real-time with moderate computation resources. 
\end{itemize}

\section{RELATED WORKS}
To reliably predict the future motion of road users, it is essential to consider the environment in which they maneuver and the interaction among them. Traditional approaches often solely consider the past and current physical states of a target road user \cite{conventional}. Deep Learning based methods have demonstrated their capability of integrating the context information into the modeling, thus reaching superior prediction performance.

\subsection{Context-awareness}
The environmental context of a scene can be present in different modalities, e.g. HD maps, semantic segmentation from perception systems, or raw data from sensors.
Different context modalities can be rasterized into bird's-eye view (BEV) grid maps and encoded with a CNN-based encoder before being fused with dynamic context features \cite{Girgis2021LatentVS,caspnet}. In \cite{deep_conv,Gilles2021,Phan-Minh2020}, road users' trajectories are also rasterized into grid maps and jointly encoded with their driving environment context using CNN-based models to learn interactions with the dynamic and static context of a scene. In \cite{gilles2021gohome}, a graph neural network (GNN) is used to learn interactions among road users. 
Different forms of attention can be used to model interactions between road users and their surroundings \cite{Graph-attention,Vectornet,Gilles2021,Girgis2021LatentVS,park2023leveraging}. In \cite{SocialLSTM,SocialGAN}, pooling operations are used to model interactions among pedestrians.

\subsection{Multi-Modal Prediction}
There can be different plausible future modes for other road users due to the unobservability of their intentions. For example, an oncoming vehicle at an intersection could drive straight or turn left and potentially cross the ego's path. Thus, planning a safe trajectory requires considering multiple plausible future modalities of other road users.
Gaussian Mixture Models (GMM) are commonly used to model multi-modal trajectories using Gaussian distributions. In \cite{Chai2019}, future trajectories are clustered to a number of mean trajectories, called anchors. Anchors' probabilities, offsets to each anchor trajectory, and covariance for the offsets are predicted. Anchors can also be defined as goals sampled from a road graph, like in \cite{Zhao2020}. Teacher forcing can be used to condition predictions to specific maneuver labels during training \cite{Deo2018_soical, sriram2020smart}. Through conditional decoding, at inference, multiple modes can be predicted.

The aforementioned methods restrict the possible future modalities to a predefined number of modes. Generative adversarial networks (GAN) \cite{GAN} and variational autoencoders (VAE) \cite{VAE} are used to facilitate multi-modality by learning to map different modalities to a continuous distribution. A best-of-N training strategy is needed \cite{SocialGAN, Bhattacharyya2018AccurateAD}. Each forward pass introduces a random sample from the distribution to predict a trajectory. The best trajectory, according to an arbitrary metric, is used for back-propagation.
However, they require multiple forward passes during training and inference and are prone to mode collapse.

\subsection{Grid-based Prediction}
Grid-based prediction approaches model possible future positions of road users using occupancy grid maps. As shown in \cite{Gilles2021, gilles2021gohome, caspnet}, multi-modality becomes an inherent property without the need for a GMM, GAN, or VAE. At the same time, grids enable the simultaneous prediction of a variable number of road users \cite{Mahjourian2022OccupancyFF, Kamenev2021PredictionNetRJ}. 
However, this comes at the cost of losing track instances which could be a drawback for specific applications.

Depending on the application, additional outputs and post-processing are necessary to extract trajectories or recover track IDs. In \cite{Mahjourian2022OccupancyFF}, a backward flow is predicted additionally for each pixel to recover track IDs, which further supports the extraction of multiple future positions for an agent. Nevertheless, it does not ensure that a track can be recovered for each agent of interest. In \cite{Kamenev2021PredictionNetRJ}, besides future occupancy, forward velocities and backtrace vectors are predicted to reconstruct trajectories starting from the current position of an agent. While starting at the initial position ensures that a track can be reconstructed for every agent of interest, it can limit multi-modality due to not reaching every predicted mode.

Another option, is to separate the prediction of multiple agents into different grid maps.
In \cite{Gilles2021, gilles2021gohome, caspnet}, goals are sampled from the prediction of the last time step, which are then used to extract trajectories.

In our approach, we combine a grid-based output for predicting all road users in a scene with an instance-based output that enables us to extract multi-modal future trajectories for any road users of interest.

\section{OUR APPROACH}

\begin{figure*}[ht!]
	\centering
	\includegraphics[width=0.79\textwidth]{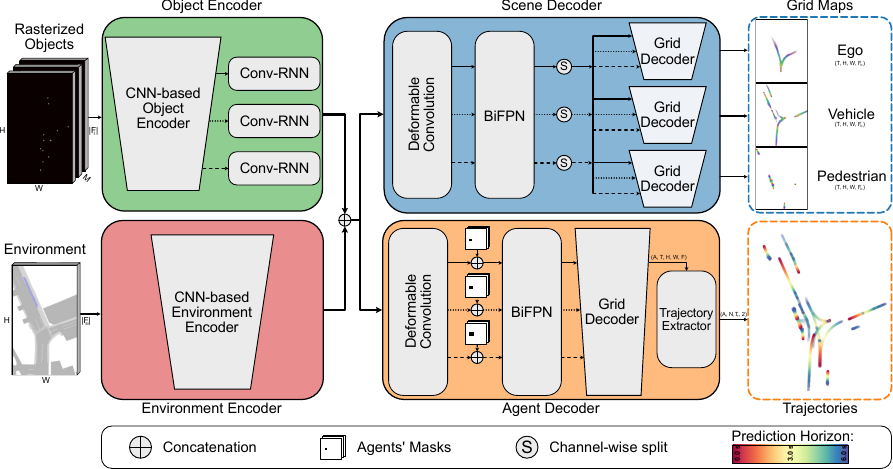}
	\caption{An architecture overview of CASPNet++. Rasterized objects are encoded and temporally fused in the Object Encoder in a pyramid structure. In the Environment Encoder, the rasterized environment surrounding the ego vehicle is encoded by a CNN-based feature pyramid network. The encoded feature maps from the Object Encoder and the Environment Encoder are concatenated on all pyramid levels and then processed in the Scene Decoder and Agent Decoder to predict the future occupancy of all road users in a scene as well as predict multi-modal trajectories for agents of interest. }
	\label{fig_architecture_detail}
\end{figure*}

\subsection{Input and Output}

\subsubsection{Object Grid}
The agents' trajectories in a scene from one or more sensors are rasterized into a series of BEV grid maps covering $M$ past time steps:
\begin{equation}
	I \in \mathbb{R}^{M \times H \times W \times |F_i|},
	\label{eq_input_structure}
\end{equation}
where $H$ and $W$ define the region size, and $|F_i|$ the number of input features. All input and output grid maps are aligned with the ego vehicle position and heading at the current time step $t_0$. The input features $F_i$ from all agents' trajectories at $t_i$ are stored in the channel dimension of an input grid and rasterized relative to the ego vehicle's position at $t_0$:
\begin{equation}
	F_i=(c^{t_i}, \delta^{t_i}, \textit{v}^{t_i}, \textit{a}^{t_i}, \sigma^{t_i}, \textit{s}^{t_i}),
	\label{eq_input_features}
\end{equation}
where $c^{t_i}$ is an one hot encoding of the object type, $\delta^{t_i}$ the in-pixel offset of an agent's position, $\textit{v}^{t_i}$ the velocity, $\textit{a}^{t_i}$ the acceleration, $\sigma^{t_i}$ the heading orientation, and $\textit{s}^{t_i}$ the bounding box size for an agent at time step $t_i$.

\subsubsection{Environment Grid}
A grid structure is chosen to represent the environment of a scene which allows an efficient fusion of different environmental context sources with a Convolutional Neural Network (CNN). Environmental context information provided by e.g. HD maps, Lidar segmentation, or Radar detection is rasterized into BEV grid maps covering the same region size as the object grids. Instead of rasterizing HD maps into a 3-channel image \cite{Djuric2020, Phan-Minh2020, caspnet}, each semantic type is represented in its own channel. Therefore, the rasterization does not lead to occlusion due to overlaps. The centerline orientations are encoded in two additional channels through the sine and cosine of the orientation at one position. Radar and Lidar detection are aggregated for the past second before the rasterization to create a denser context grid map.

\subsubsection{Output}
The output $O_S$ of the Scene Decoder is defined through multiple grids for $T$ future time steps and $C$ object types, describing the future occupancy probabilities for all objects from one object type as well as an in-pixel offsets and velocities:
\begin{equation}
	O_S \in \mathbb{R}^{C \times T \times H \times W \times |F_o|}.
	\label{eq_output_scene}
\end{equation}
The output features $F_0$ per pixel are defined as follows:
\begin{equation}
	F_o=(o^{t_o}, \delta^{t_o},  \textit{v}^{t_o}),
	\label{eq_output_features}
\end{equation}
where $o^{t_o}$ describes the future occupancy for an object type or agent, $\delta^{t_o}$ the in-pixel offset for predicted pixel in $h$ and $w$ direction, and $\textit{v}^{t_o}$ future velocities in $h$ and $w$ at time $t_o$.
The fixed input and output structure of the Scene Decoder enables an efficient prediction of all road users in a scene where the computational cost is independent of the number of road users and the complexity of a scenario.

However, for specific applications, such as trajectory planning, it can be necessary to have an instance-based output where multiple futures are separable. Therefore, we developed an Agent Decoder, which predicts agents' future trajectories based on the same input feature maps as the Scene Decoder. The intermediate output of the Agent Decoder is a series of grid maps similar like $O_S$:
\begin{equation}
	O_A \in \mathbb{R}^{A \times T \times H \times W \times |F_o|},
	\label{eq_output_agent}
\end{equation}
where $A$ describes the number of agents of interest. $O_A$ is processed in the Trajectory Extractor to reconstruct $N$ future trajectories, consisting of a series of $h, w$ positions, for the $A$ agents:
\begin{equation}
	O_T \in \mathbb{R}^{A \times N \times T_T \times 2},
	\label{eq_output_trajectories}
\end{equation}
where $T_T$ are the number of time steps.

\subsection{Network Architecture}
In comparison to our previous work, the focus of this work was to further utilize the advantages of grid-based input and output structures. At the same time, our goal was to enable an instance-based agent prediction and improve the overall performance through enhanced interaction modeling.

Throughout the architecture, a CNN-based Feature Pyramid Network (FPN) structure is used to capture agents' interactions with their surroundings at different ranges on the different pyramid levels. By introducing Deformable convolution \cite{Zhu2018DeformableCV} in the decoders, a flexible kernel coverage can be achieved for each pixel, allowing the network to efficiently model interactions among road users at a higher range without increasing the number of pyramid levels.
To learn more complex interactions by fusing multi-scale interaction features earlier in the network, a BiFPN \cite{Tan2019EfficientDetSA} is utilized after the Deformable convolution layers. CASPNet's Grid Decoder \cite{caspnet} processes the feature maps from all pyramid levels to predict future occupancy probabilities, in-pixel offsets, and velocities. 

An overview of the architecture is depicted in Fig. \ref{fig_architecture_detail}. CASPNet++ consists of four key components: The Object Encoder, Environment Encoder, Scene Decoder, and Agent Decoder. 

\subsubsection{Object Encoder}
A CNN-based FPN encodes the objects frame per frame. The resulting feature maps for the $M$ input time steps are temporally fused on every level through Convolutional-RNN layers.

\subsubsection{Environment Encoder}
Due to the prediction of different object types (e.g. pedestrian and vehicle) that maneuver differently and interact with their environment at different ranges, the environmental input grids are processed by a CNN-based FPN capturing multi-scale features. The feature maps are concatenated on every pyramid level with the same resolution feature maps from the Object Encoder and passed to the Scene Decoder and the Agent Decoder.

\subsubsection{Scene Decoder}
The Scene Decoder fuses the object and environmental features spatially on every pyramid level as well as over multiple scales through the BiFPN. After the BiFPN, the feature maps are split at the channel dimension at every level and processed by separate Grid Decoders to predict $O_S$ for $C$ object types, such as the ego vehicle, other vehicle, and pedestrian.

\subsubsection{Agent Decoder}
To reduce computational costs and facilitate real-time capability, the feature maps from the encoders are shared for the Agent and Scene Decoder.
After applying Deformable Convolution, the resulting feature maps are extended to one additional dimension and repeated $A$ times to cover each agent instance. As a conditional input to indicate which agents should be predicted, agent masks are used. An agent mask consists of a 2D binary representation of the agent's position at $t_0$, rasterized according to a pyramid's level resolution. There are two additional channels for each agent's mask where an in-pixel offset describes the exact position of the center point of an agent at $t_0$. The agents' masks are concatenated with the feature maps and then processed by the BiFPN and a Grid Decoder to produce the intermediate grid-based output $O_A$. The Trajectory Extractor processes $O_A$ to create the final trajectory output for $A$ agents.

\subsection{Trajectory Extractor}
In order to reconstruct trajectories for an agent $a$ from the intermediate Agent Decoder output $O_A^a$, we first get the prediction horizon through the grid with the most points over a threshold corresponding to the agent. Subsequently, $N$ goals are sampled from the corresponding grid map. We use a MLP conditioned on the agent's current state, the goals, and the prediction horizon to predict initial future trajectories. The initial trajectories are then refined by the agent's output grid maps  $O_A^a$ as described in our previous work \cite{caspnet} to produce $O_T$.

\subsection{Loss Function}
As Ground Truth (GT) $Y$, we rasterize the future positions of the road users separately for each object type and agent of interest, resulting in a series of 2D grid maps. Around GT positions, a Gaussian Kernel is added to reduce the loss in the vicinity (see Eq. \ref{eq_class_loss}). GT velocities and in-pixel offsets are included in the channel dimension.

Alike our previous work, we use a distance-aware focal loss for the future occupancy prediction task, inspired by \cite{Law_2018_ECCV} and \cite{focalLoss}:
\begin{equation}
	\resizebox{0.91\hsize}{!}{%
		$	{L_{ct}^{\text{occupancy}} = {-1 \over {N_{ct}+1}} \sum_{h=1}^{H} \sum_{w=1}^{W}
			\begin{cases}
				\alpha (1-p)^\gamma log(p) & \text{if $Y_{ctwh}$ = 1}\\
				(1-\alpha) (1-Y_{twhc})^\beta p^\gamma log(1-p) & \text{otherwise}
		\end{cases}}$%
	}
	\label{eq_class_loss}
\end{equation}
where p is the predicted occupancy for object type $c$ or agent $a$ at time step $t$. $N_{ct}$ describes the number of objects in $Y_{ct}$.

For the in-pixel offset and the velocity prediction task, we use the L2 loss, which is evaluated only for pixel positions where an object is located in the GT occupancy:
\begin{equation}
	\resizebox{0.81\hsize}{!}{%
		$	{
			L_{ct}^{f} = w_{f} \sum_{h=1}^{H} \sum_{w=1}^{W}
			\begin{cases}
				(O_{thwf_{h}} - Y_{thwf_{h}})^2 + \\ (O_{thwf_{w}} - Y_{thwf_{w}})^2 & \text{if $ \sum_{c=1}^{C} Y_{cthw} = 1 $}\\
				0 & \text{otherwise}
		\end{cases}}$,%
	}
	\label{eq_feature_loss}
\end{equation}
where $w_f$ is a weight for task $f \in \{\text{offset}, \text{velocity} \}$.

The loss for the scene prediction is defined as follows:
\begin{equation}
	\resizebox{0.76\hsize}{!}{%
		$	{ L_{\text{Scene}} = \sum_{c=1}^{C} \sum_{t=1}^{T} w_{c} (L_{ct}^{\text{occupancy}} +  L_{ct}^{\text{offset}} + L_{ct}^{\text{velocity}}),
		}$%
	}
	\label{eq_scene_loss}
\end{equation}
where $w_c$ is a weight for object type $c$. The loss for the agents is defined accordingly:
\begin{equation}
	\resizebox{0.76\hsize}{!}{%
		$	{L_{\text{Agents}} = w_A \sum_{a=1}^{A} \sum_{t=1}^{T} (L_{at}^{\text{occupancy}} +  L_{at}^{\text{offset}} + L_{at}^{\text{velocity}}).
		}$%
	}
	\label{eq_agent_loss}
\end{equation}
The final loss is given as the sum over the agent and scene loss divided by the number of predicted time steps $T$:
\begin{equation}
	\resizebox{0.41\hsize}{!}{%
	 $	{Loss =  {1 \over T}(L_{\text{Scene}} + L_{\text{Agents}}).
		}$%
    }
	\label{eq_joint_loss}
\end{equation}

\begin{figure}[ht!]
	\centering
	\includegraphics[width=0.49\textwidth]{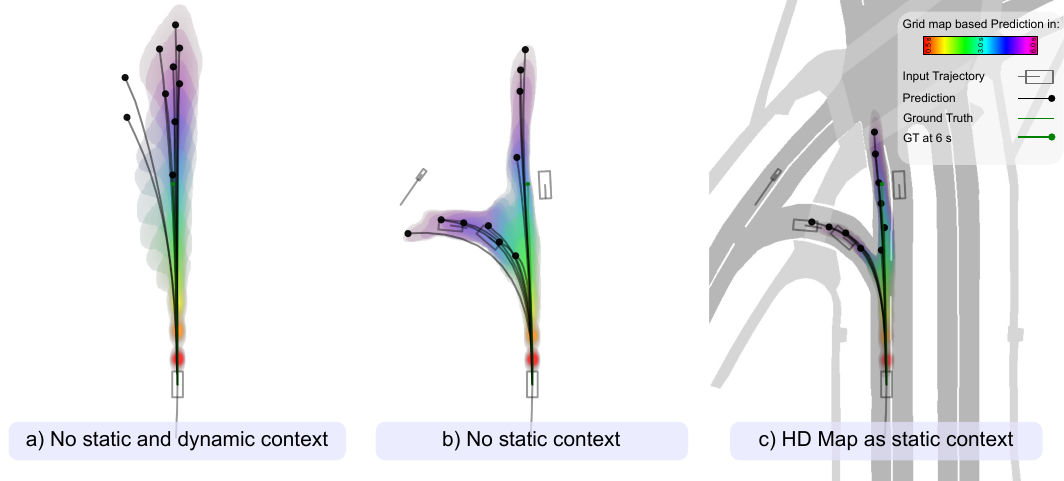}
	\caption{Qualitative examples of CASPNet++ using different inputs to predict the same target. The grid-based output is visualized with a color map, where red indicates the prediction in 0.5 s and magenta the prediction in 6 s. The black lines illustrate the predicted trajectories. In a), the prediction is only based on the current and past states of the target vehicle. In b), other objects are additionally used as input. The two turning vehicles help the network to assume a possibility of a left turn. In c), the utilization of the HD map leads to a more certain prediction.}
	\label{fig_map_ablation_a}
\end{figure}

\begin{figure}[ht!]
	\centering
	\includegraphics[width=0.49\textwidth]{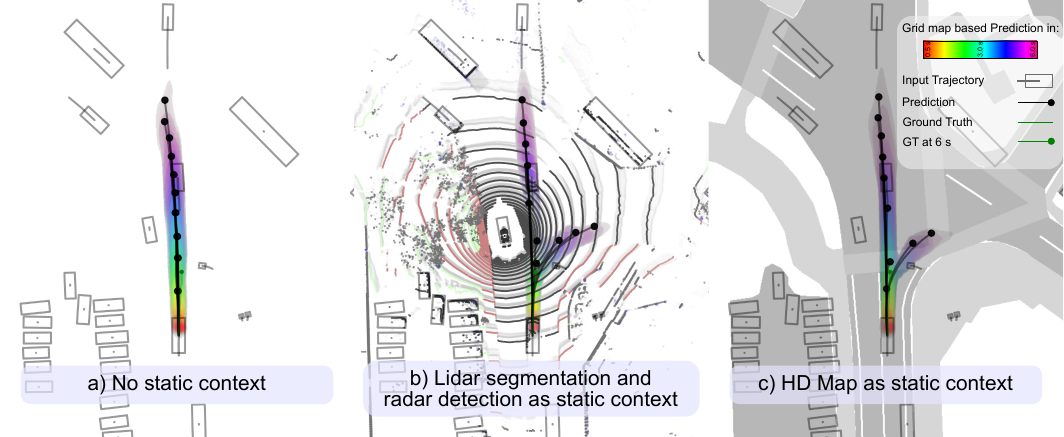}
	\caption{Qualitative examples of CASPNet++ using different environmental input modalities to predict the same target. In a), the model predicts without environmental input. In b), Lidar segmentation and Radar detection are processed by the model.  In c), the HD map is used as environmental input.}
	\label{fig_map_ablation_b}
\end{figure}

\begin{figure*}[ht!]
	\centering
	\includegraphics[width=0.99\textwidth]{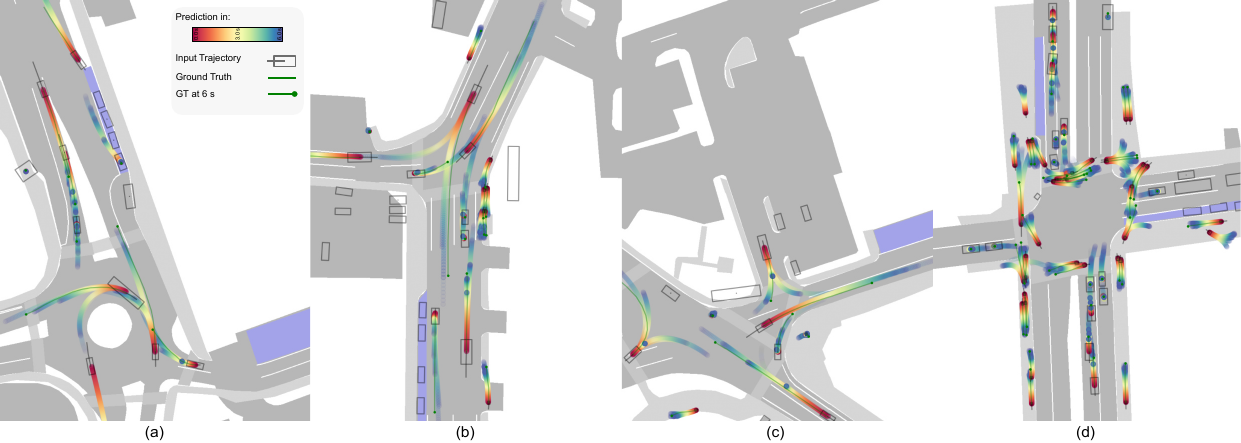}
	\caption{Qualitative examples of CASPNet++ multi-agent trajectory prediction on the nuScenes dataset. The time horizon of the predicted trajectories is visualized as a heatmap from red (the closest time step) to blue (the time step in 6 s). The probability of each trajectory is visualized through the transparency. The black bounding boxes represent the current position of road users, and the black lines represent their past trajectories. The ground truth trajectories are shown in green.}
	\label{fig_qualitative}
\end{figure*}

\subsection{Implementation Details}
\subsubsection{Network Architecture, Input and Output Definition}
Four pyramid levels are used throughout the network architecture, with (16, 32, 64, and 128) feature output channels, respectively. The CNN blocks consist of 2D convolution, batch normalization, ReLU layers, dropout, and Max Pooling. After the Deformable convolution, the feature maps on all pyramid levels have 64 channels.

For nuScenes and the in-house dataset, the input and output resolution is 1.0 m per pixel and the region size is $H=152$, and $W=96$ where the target or ego vehicle is located at pixel position ($h=122$, $w=48$). The number of past time steps is 3 with a frame rate of 2 Hz. 12 time steps are predicted with a frame rate of 2 Hz. For nuScenes and INTERACTION, we used the target vehicle as the ego vehicle. 
Accordingly, we center the input and output grid maps around the target vehicle to make sure that the region fully covers the future maneuverability. 
Due to the smaller prediction horizon of 3 s and therefore, a smaller prediction region, we increased the resolution to 0.5 m per pixel for INTERACTION.

\subsubsection{Training, Data Augmentation and Inference}
The models were trained on one Nvidia GeForce RTX 2080Ti GPU. The batch size is 16. The AdamW optimizer \cite{adamw} with an initial learning rate of 0.001 is used. 
During training, we rotate the samples with a probability of 0.75 between [$-{\pi \over 3}$, ${\pi \over 3}$] and apply a random translation [-3, 3]. Furthermore, we randomly flip along the y-axis and randomly drop agents other than the target.
We set $w_\text{offset} = w_\text{velocity} = 0.1$, $(0.25, 0.25, 0.1)$ for the object types (ego, vehicle, pedestrian), and $w_A=0.4$. 

For online running, CASPNet++ is deployed on an Nvidia GeForce RTX 2080 together with a machine learning based perception system, running in real-time at 20 Hz.

\section{EVALUATION}\label{evaluation}
\subsection{Datasets}
For evaluating our proposed method, we use two popular datasets in motion prediction: nuScenes \cite{Caesar2020} and INTERACTION \cite{zhan2019interaction}. In addition, an in-house dataset is used for an ablation study on the network architecture.

\subsubsection{nuScenes}
nuScenes \cite{Caesar2020} consists of 1000 scenes of 20 s length recorded in urban areas in Boston and Singapore. Given a maximum of 2 s of a target agent's history, the goal is to predict the future trajectory for the next 6 s. For our experiments, we only used 1 s of history. HD maps are provided to give context about the static road environment.

\subsubsection{INTERACTION}
INTERACTION \cite{zhan2019interaction} is a drone-based recorded dataset that includes highly interactive scenarios such as roundabouts, un-signalized intersections, and lane merging. 1 s of agents' history and an HD map are provided to predict the next 3 s of a target agent. 

\subsubsection{In-house Dataset}
Our in-house dataset for motion prediction consists of 120 hours of interactive urban-focused traffic scenarios such as intersections, roundabouts, and narrow roads recorded in Las Vegas, Krakow, Wuppertal, and Düsseldorf. Maps for all regions are generated by trail map aggregation \cite{map_aggregation}. Objects' trajectories are provided by a Radar-based perception system. 1 s of history is used to predict the next 6 s.

\subsection{Metrics}
We apply the most widely used metrics in motion prediction: The Minimum Final Displacement Error ($minFDE_k$)  minimum Average Displacement Error ($minADE_k$) and Miss Rate ($MR_k$). The metrics are calculated for different number of future modalities $k$.
For nuScenes, we additionally report the off-road rate ($OR$), which describes the ratio of predicted trajectories outside the drivable region to the total number of predictions.

\subsection{Comparison with state-of-the-Art}
We compare the result of CASPNet++ on the nuScenes prediction challenge to the state-of-the-art methods in Tab. \ref{tab_nuscenes_result}. In order to allow a fair comparison with other approaches, we additionally demonstrate ensemble model results. The output grid maps of two independently trained CASPNet++ models are averaged, resulting in the CASPNet++ Ensemble model. We achieve state-of-the-art results for ${minADE_5}$, $minFDE_1$, $minADE_1$, $OR$ and $MR_{10}$.

\begin{table*}[ht]
	\caption{Results on the nuScenes prediction challenge test split.}
	\centering
	\resizebox{0.69\hsize}{!}
	{%
		\begin{tabular}{ l | r r | r r | r r  | r}
			\textit{} & \multicolumn{2}{c} {k=10} & \multicolumn{2}{c} {k=5}  & \multicolumn{2}{c} {k=1} & \textit{} \\
			
			Method & \textit{MR} $\downarrow$ & \scriptsize \textit{min}ADE $\downarrow$ & \textit{MR} $\downarrow$ & \scriptsize \textit{min}ADE $\downarrow$ & \scriptsize \textit{min}FDE $\downarrow$ & \scriptsize \textit{min}ADE $\downarrow$ & \textit{OR} $\downarrow$
			\\  
			\hline
			Constant Velocity \cite{Phan-Minh2020} & 0.91 & 4.61 & 0.91 & 4.61 & 11.21 & 4.61 & 0.14 \\
			Physics Oracle \cite{Phan-Minh2020} & 0.88 & 3.70 & 0.88 & 3.70 & 9.09 & 3.70 & 0.12 \\
			CoverNet \cite{Phan-Minh2020} & 0.64 & 1.92 & 0.76 & 2.62 & 11.36 & - & 0.13 \\
			\hline
			WIMP \cite{Khandelwal2020} & 0.43 & 1.11 & 0.55 & 1.84 & 8.49 & - & 0.04 \\
			MHA-JAM \cite{Messaoud2020} & 0.45 & 1.24 & 0.59 & 1.81 & 8.57 & 3.69 & 0.07 \\
			P2T \cite{deo2021trajectory} & 0.46 & 1.16 & 0.64 & 1.45 & 10.50 & - & 0.03 \\
			GOHOME \cite{gilles2021gohome} & 0.47 & 1.15 & 0.57 & 1.42 & 6.99 & - & 0.04 \\
			CASPNet \cite{caspnet} & 0.43 & 1.19 & 0.60 & 1.41 & 7.27 & 3.16 & 0.02 \\
			Autobot \cite{Girgis2021LatentVS} & 0.44 & 1.03 & 0.62 & 1.37 & 8.19 & - & 0.02 \\
			PGP \cite{Deo2021MultimodalTP} & 0.34 & 0.94 & 0.52 & 1.27 & 7.17 & - & 0.03 \\
			FRM \cite{park2023leveraging} & 0.30 & \textbf{0.88} & \textbf{0.48} & 1.18 & 6.59 & - & 0.02 \\
			\hline
			CASPNet++ & 0.30 & 0.93 & 0.50 & 1.18 & 6.19 & \textbf{2.74} & \textbf{0.01} \\
			\scriptsize CASPNet++ Ensemble & \textbf{0.29} & 0.92 & 0.50 & \textbf{1.16} & \textbf{6.18} & \textbf{2.74} & \textbf{0.01} \\
	\end{tabular}}
	\label{tab_nuscenes_result}
\end{table*}

\subsection{Ablation Studies}
\subsubsection{Ablation study on different Environmental Input Modalities}
To evaluate the capability of CASPNet++ to cope with diverse environmental input modalities, we additionally trained and tested our model on Lidar segmentation and Radar detection as input. The results are shown in Tab. \ref{tab_map_ablation}.

When only using the dynamic context, the model can often make reasonable predictions based on other vehicles' movements. For example, in Fig. \ref{fig_map_ablation_a} (b), $CASPNet\text{++}$ predicts that the target can turn left based on two leading turning vehicles which is not the case for the model using only the target's past states, in Fig. \ref{fig_map_ablation_a} (a). However, when compared with the model using an HD map, the uncertainty is much higher, in Fig. \ref{fig_map_ablation_a} (c). 
In Fig. \ref{fig_map_ablation_b} (a), $CASPNet\text{++}$ fails to predict the availability of a right turn solely based on the dynamic input. When including Lidar segmentation and Radar detection, the possibility of a right turn is predicted, in Fig. \ref{fig_map_ablation_b} (b).

\begin{table}[ht]
	\caption{Ablation study on different environmental context input modalities (nuScenes validation split).}
	\centering
	\resizebox{0.99\hsize}{!}
	{%
		\begin{tabular}{l | c | c | c | r r | r r | r }
			\hline
			\textit{} & \scriptsize HD & \scriptsize Lidar & \scriptsize Radar & \multicolumn{2}{c} {k=5}  & \multicolumn{2}{c} {k=1} \\
			
			& \scriptsize Map & \scriptsize Seg. & \scriptsize Det. & \scriptsize \textit{min}ADE $\downarrow$ & \textit{MR} $\downarrow$ & \scriptsize \textit{min}FDE $\downarrow$ & \textit{MR} $\downarrow$ & \textit{OR} $\downarrow$ 
			\\  
			\hline
			Ego & &  &  & 1.41 & 0.55 & 6.98 & 0.79 & 0.17 \\
			& $\checkmark$ & & & \textbf{1.02} & 0.39 & \textbf{5.31} & \textbf{0.69} & 0.02 \\
			& & $\checkmark$ & & 1.19 & 0.45 & 6.46 & 0.77 & 0.04 \\
			& &  & $\checkmark$ & 1.24 & 0.47 & 6.24 & 0.74 & 0.07 \\
			& & $\checkmark$ & $\checkmark$ & 1.16 & 0.46 & 6.17 & 0.74 & 0.04 \\	
			& $\checkmark$ & $\checkmark$ & $\checkmark$ & 1.06 & \textbf{0.37} & 5.32 & \textbf{0.69} & \textbf{0.01} \\
			\hline
			Target & &  &  & 1.46 & 0.59 & 7.27 & 0.82 & 0.12 \\
			& $\checkmark$ &  &  & \textbf{1.17} & 0.49 & \textbf{6.19} & \textbf{0.79} & \textbf{0.01} \\
			& & $\checkmark$ & & 1.38 & 0.53 & 6.94 & 0.82 & 0.05 \\
			& &  & $\checkmark$ & 1.39 & 0.56 & 7.04 & 0.82 & 0.09 \\
			& & $\checkmark$ & $\checkmark$ & 1.38 & 0.54 & 6.99 & 0.82 & 0.05 \\			
			& $\checkmark$ & $\checkmark$ & $\checkmark$ & \textbf{1.17} & \textbf{0.48} & 6.25 & \textbf{0.79} & \textbf{0.01} \\
			\hline
	\end{tabular}}
	\label{tab_map_ablation}
\end{table}

Since we use perception-based input of the environment, the information density of the surroundings differs for the ego vehicle and other vehicles. Thus, we divided the evaluation into ego and target vehicles.
Tab. \ref{tab_map_ablation} shows that the perception-based models perform better than the model only using the dynamic context for ego as well as target prediction. The OR for the target prediction decreases from 12 \% to 5 \%, indicating an improved understanding of the road structures. 
For the ego prediction the improvement is even higher (17 \% to 4 \%).

When comparing the perception-based models with the model using the HD map, it shows the perception-based models perform worse. One of the reasons, might be that information about exact lane location and driving directions, which are included in HD maps, are not fully observable from Lidar and Radar. Vision-based lane boundaries as additional input could help to overcome this challenge in future experiments.

Moreover, these experiments showed that grids are a suitable structure to represent different environmental context sources in a joint format to enable an efficient fusion using CNN in CASPNet++'s Environment Encoder. 

\subsubsection{Network Architecture Ablation Study}
CASPNet \cite{caspnet} represents the baseline model in the ablation study on the network architecture in Tab. \ref{tab_network_ablation}. Multi-scale feature fusion through BiFPN in the decoder significantly improves the performance. 
Having flexible receptive fields for pixels can help the network to cope with objects' different maneuverabilities.
We experimented with Deformable Convolution as a replacement for the Attention Blocks \cite{caspnet}.

\begin{table}[ht!]
	\caption{Ablation study on the network architecture (In-house dataset).}
	\centering
	\resizebox{0.99\hsize}{!}
	{%
		\begin{tabular}{ l | r r | r r }
			\hline
			\textit{} & \multicolumn{2}{c} {k=1}  & \multicolumn{2}{c} {k=5} \\
			
			Method & \scriptsize \textit{min}FDE $\downarrow$ & \textit{MR} $\downarrow$ & \scriptsize \textit{min}FDE $\downarrow$ & \textit{MR} $\downarrow$ 
			\\
			\hline
			Constant Velocity & 13.99 & 0.92 & 13.99 & 0.92 \\
			CASPNet \cite{caspnet} & 6.58 & 0.78 & 2.79 & 0.46 \\
			CASPNet (BiFPN \cite{Tan2019EfficientDetSA}) & 5.84 & 0.77 & 2.41 & 0.40 \\
			CASPNet (Deformable Conv. \cite{Zhu2018DeformableCV}) & 5.87 & \textbf{0.74} & 2.36 & 0.42 \\
			\hline
			CASPNet++ (Transformer-based Encoder \cite{Lee2018SetTA}) & 6.24 & 0.77 & 2.49 & 0.43 \\
			CASPNet++ & \textbf{5.74} & 0.75 & \textbf{2.32} & \textbf{0.40} \\
			\hline
	\end{tabular}}
	\label{tab_network_ablation}
\end{table}

The combination of the BiFPN and Deformable Conv. model is denoted as CASPNet++ and is depicted in Fig. \ref{fig_architecture_detail}. The combined model outperforms all others and shows significant improvement over the predecessor model $\text{CASPNet}$.
We evaluated a Set-Transformer-based \cite{Lee2018SetTA} Object Encoder inspired by \cite{Girgis2021LatentVS}, to learn interactions on the trajectory level where far-range interactions are not limited through the receptive field of a CNN. However, in our experiments, it did not improve the performance. We plan to further investigate Transformer-based architectures for motion prediction in future works.

\subsubsection{INTERACTION}
We show quantitative results on the INTERACTION dataset in Tab. \ref{tab_interaction_result}. Due to the unavailability of the evaluation server at the time of the submission, we had to use the validation split. For the evaluation, we considered every vehicle that has at least ten input time steps and 30 GT time steps as a target.

\begin{table}[bt]
	\caption{Results on the INTERACTION dataset (validation split).}
	\centering
	\resizebox{0.99\hsize}{!}
	{%
		\begin{tabular}{ l | r r r | r r r}
			\textit{} & \multicolumn{3}{c} {k=6} & \multicolumn{3}{c} {k=1} \\
			Method & \scriptsize \textit{min}ADE $\downarrow$ & \scriptsize \textit{min}FDE $\downarrow$ & \textit{MR} $\downarrow$ &  \scriptsize \textit{min}ADE $\downarrow$ & \scriptsize \textit{min}FDE $\downarrow$ & \textit{MR} $\downarrow$ \\
			\hline
			CASPNet & 0.14 & 0.32 & 0.01 & 0.24 & 0.75 & 0.15 \\
			CASPNet++ & 0.12 & 0.28 & 0.01 & 0.22 & 0.70 & 0.13 \\
	\end{tabular}}
	\label{tab_interaction_result}
\end{table}

\subsection{Evaluation Agent-Conditioned Prediction}	
\subsubsection{Qualitative Results}
In Fig. \ref{fig_qualitative}, qualitative results for the agent-conditioned trajectory prediction are shown in four challenging scenarios: A roundabout (a), a 3-way intersection (b), an irregular intersection (c), and a 4-way intersection (d). When taking a closer look at (b), we can see that CASPNet++ makes plausible multi-modal trajectory predictions for all road users considering their respective dynamic and environmental contexts. The vehicle coming from the top can turn right, or drive straight, which has a higher probability (visualized by a higher opacity). Depending on its maneuver, the road user waiting at the intersection can directly start driving, or wait a little longer, which correspondingly has a higher probability.

\subsubsection{Quantitative Results}
\begin{figure}[h]
	\centering
	\includegraphics[width=0.49\textwidth]{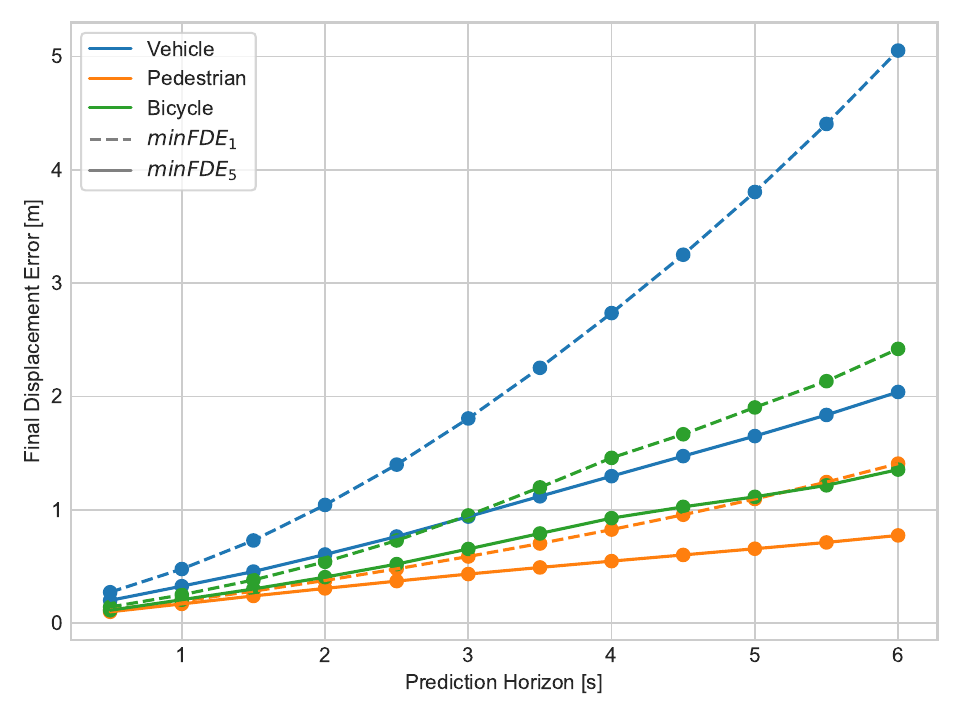}
	\caption{Quantitative results of CASPNet++ agent conditioned prediction evaluated on the nuScenes validation split. We report the FDE at multiple time horizons for $k=1$ (dashed line) and $k=5$ (solid line).
	}
	\label{fig_agent_decoding_quantitative}
\end{figure}

We show quantitative results of the agent-conditioned trajectory prediction in Fig. \ref{fig_agent_decoding_quantitative}. All road users that are in the region of interest at the current time step are predicted and evaluated. As a comparison, in Tab. \ref{tab_nuscenes_result} only target vehicles are evaluated. Parked vehicles are excluded from the metric calculation. The $minFDE_k$ for $k=(1, 5)$ for the object types vehicle, pedestrian, and bicycle are reported at multiple future time horizons. 

\section{SUMMARY AND FUTURE WORKS}
In this work, focused on utilizing the advantages of grid-based input and output structures, CASPNet++ has been designed to efficiently predict the motion of all road users in a scene. An instance-based output head has been introduced that predicts multi-modal trajectories of agents to support motion planning and ADAS features. We showed that our model can effectively fuse different environmental context sources (e.g. Radar detection and Lidar segmentation) and provide context-aware predictions without an HD map. In quantitative evaluations, we show that our model reaches state-of-the-art results on the challenging urban-focused dataset nuScenes. CASPNet++ has been tested in a prototype test vehicle where it provides reliable predictions of the scene in real-time using sensor-based perception input.

In future works, we plan to integrate vision-based perception as additional input, such as lane detection or BEV-based semantic segmentation.
As additional output, sets of scene-consistent trajectories for all road users could be extended. Moreover, we will further develop the conditional agent prediction to support decision-making and planning tasks. 

\section*{ACKNOWLEDGEMENT}
This work is a result of the joint research project STADT:up. The project is supported by the German Federal Ministry for Economic Affairs and Climate Action (BMWK), based on a decision of the German Bundestag. The author is solely responsible for the content of this publication. 


\bibliographystyle{IEEEtran}
\bibliography{./IEEEtranBST/IEEEexample}
\end{document}